# Multi-Sensor Image Fusion Based on Moment Calculation

Sourav Pramanik
Computer Science and Engineering Department
National Institute of Science and Technology
Berhampur, India
srv.pramanik03327@gmail.com

Debotosh Bhattacharjee
Computer Science and Engineering Department
Jadavpur University
Kolkata, India
debotosh@ieee.org

*Abstract*— **An image fusion method based on salient features is proposed in this paper. In this work, we have concentrated on salient features of the image for fusion in order to preserve all relevant information contained in the input images and tried to enhance the contrast in fused image and also suppressed noise to a maximum extent. In our system, first we have applied a mask on two input images in order to conserve the high frequency information along with some low frequency information and stifle noise to a maximum extent. Thereafter, for identification of salience features from sources images, a local moment is computed in the neighborhood of a coefficient. Finally, a decision map is generated based on local moment in order to get the fused image. To verify our proposed algorithm, we have tested it on 120 sensor image pairs collected from Manchester University UK database. The experimental results show that the proposed method can provide superior fused image in terms of several quantitative fusion evaluation index.**

## I. INTRODUCTION

In recent years, multi-sensor image fusion has received weighty attention for both military and industrial applications. For detailed assessment, it is always not possible to have all the physical and geometrical information in a single image. To achieve this, image fusion is call for. Image fusion is defined as the process of combining two or more different images into a new single image retaining important features from each image and the quality of the fused image is superior to any of the input images. According to the stage at which the information is combined, image fusion algorithms can be categorized into three levels, namely pixel, feature and decision level [1]. Image fusion at pixel level means fusion at the lowest processing level referring to the merging of measured physical parameters. Feature level fusion requires first extraction of the features, those features can be identified by characteristics such as contrast, size, shape and texture. Symbol level fusion allows the information to be effectively combined at the highest level of abstraction. The selection of the convenient level depends on many factors such as data sources, application and available tools.

In recent years, many techniques for generic image fusion have been designed. Among this techniques, there are lots of methods have been proposed on pixel level
image fusion. The simplest image fusion on pixel level is to sum and average the original images pixel by pixel. However when this method is applied, several undesired effects including reduction in contrast of feature would appear. In [2] recognized that multi-scale transforms are very useful for analyzing the information content of images for the purpose of fusion. Typical multi-scale transforms include the Laplacian pyramid [3], morphological pyramid [4], discrete wavelet transform (DWT) [5-7], gradient pyramid [8], stationary wavelet transform (SWT) [9], [10], and dual-tree complex wavelet transform (DTCWT) [11], [12]. Recently developed multiscale geometry analysis, such as ridgelet transform [13], Curvelet transform (CVT) [14], the nonsubsampled contourlet transform (NSCT) [15], [16], are also applied to image fusion. Apart from pixel by pixel averaging and multi-scale transform, few techniques has been exploited on feature level fusion. In [17], the salient features are first identified in each source image then based on salient features a selection mode is employed. In [18], the high frequency coefficient is obtained by choosing the corresponding coefficient with the greater local deviation. In [19], images are fused based on edges.

In this paper, we have focused on the feature level image fusion problem. The goal of this fusion is to combine visual information contained in multiple source images into an informative fused image without the introduction of distortion or loss of information. In this work, we have concentrated on salient features of the image for fusion in order to preserve all relevant information contained in the input images and tried to enhance the contrast in fused image and also suppressed noise to a maximum extent. In our system, first we have applied a mask on two input images in order to conserve the high frequency information along with some low frequency information and stifle noise to a maximum extent. Thereafter, for identification of salience features from sources images, a local moment is computed in the neighborhood of a coefficient. Finally a decision map is generated based on local moment in order to get the fused image. To verify our proposed algorithm, we have tested it on different sensor images collected from Manchester University UK database. The rest of the paper is organized as follows: the section II describes the overall system design and details of the experiments conducted along with results are given in section III, section IV concludes the paper.

## II. PROPOSED METHOD

In this paper, we have employed a feature based image fusion method. The main aims of image fusion are fused image must preserve as much as possible all the relevant information that are present in the input images and the fusion process should not introduce any inconsistencies or irrelevant information that can distract or mislead the human observer or any subsequent processing steps. Apart from that, in fused image, noise should be suppressed to a maximum extent. In response to all the requirements, we have considered salient features of the input images, which must be preserved in the fused image. The underlying assumption of our method is that salient features are the features which consists most of the important information (such as edges, contrast, etc) of an image because we know that useful features in the image usually are larger than its neighbor pixel. The main problem for image analyses is that noise and irrelevant information. For removal of noise, we have used a high pass filter mask and for removal of irrelevant information and to conserve important information in the fused image, a local moment is computed in the neighborhood of a coefficient. This section is divided into two parts: Preprocessing, Moment calculation and Fusion.

### A. Preprocessing

In this section, we have done some preprocessing task on input images before fusion. The main problem of image analysis is noise or irrelevant information. Before fusion of two or more images into a single image, we have to make sure that source images are noise free to a maximum extent and images should content high frequency information. To achieve this, we have used here a filter mask which shown in fig.1. In this filter mask, we have considered center value 17.9 because to preserve high frequency information along with low frequency information in the image and to remove noise to a maximum extent.

### B. Moment Calculation and Fusion

In this section, we have calculated moment of a coefficient. In order to preserve the salient features (like edges, corner, brightness, etc) in the fused image, a moment is calculated. Our assumption is that, the moment of a coefficient in the neighborhood signifies how much information it contains. So the salience of a feature is computed here as a local moment in the neighborhood of a coefficient. We computed here local moment because to avoid slow computational speed up. The local moment is calculated using eq.1 [20].

$$M(X, a) = \sum_{c=1}^{3}\sum_{r=1}^{3} r^i c^j\, I(X, b) \qquad \ldots\ldots\ldots\ldots\ldots\ldots (1)$$

Where, $M(X, a)$ is the moment of source image $X$ at point $a$. The closer the point $b$ is near the point $a$ and $i, j$ are the coordinates of the point $a$. Here, the neighborhood window is $3 \times 3$ centered at the current coefficient position. $M(Y, a)$ is also computed by the same rule. Finally, a decision map is

$$\frac{1}{9} \times \begin{pmatrix} -1 & -1 & -1 \\ -1 & 17.9 & -1 \\ -1 & -1 & -1 \end{pmatrix}$$

Fig.1 Filter mask

implemented based on calculated moment to achieve a fused image from two source images which preserve all the relevant information those are present in the source images. The decision map is implemented using the eq. 2.

$$F(r, c) = \begin{cases} X(r, c)\ if\ M(X, a) \geq M(Y, a) \\ Y(r, c)\ if\ M(Y, a) > M(X, a) \end{cases} \ldots\ldots\ldots\ldots\ldots (2)$$

Where, $F(r, c)$ is a fused image corresponding to two source images.

## III. EXPERIMENTAL RESULTS

This section shows the experimental results in terms of performance index. Here we have tested our algorithm on 120 pairs of sensor images collected from Manchester University UK database without reference images. The performance evaluation is done by computing Mutual Information Measure (MIM) which measures the degree of dependence of the two images and a larger measure implies better quality [21], Standard Deviation (SD) which measures the contrast in the fused image and an image with high contrast would have a high standard deviation [21] and Entropy which measure the information content of an image and an image with high information content would have high entropy [22]. We compare our proposed algorithm with three other well known fusion methods. The value of each quality assessment metrics of four fusion approaches on Fig.2, Fig.3, and Fig.4 are given in table 1. The experimental result shows that our proposed algorithm gives better performance and quality in comparison with conventional wavelet transform [22], principal component analysis [21] and weighted average discrete wavelet transform using genetic algorithm [22].

Finally, we have compared our method with Petrovic method [19] in terms of objective evaluation metric $Q_P^{AB/F}$ [23]. An objective fusion measure should i) extract all the perceptually important information that exists in the input images and ii) measure the ability of the fusion process to transfer as accurately as possible this information into the output image [23]. By this metric, we can evaluate the amount of edge information that is transferred from input images to the fused image. The value of $Q_P^{AB/F}$ vary in between 0 and 1. A value of 0 corresponds to the complete loss of edge information at location $(n, m)$ and a value of 1 indicates no loss of information. The $Q_P^{AB/F}$ is calculated using eq.3 [23]. In this paper, we have used this evolution technique to evaluate performance of our proposed method. The average $Q_P^{AB/F}$ results over the 120 input pairs collected from Manchester University UK database, of our scheme and the DWT fusion with XOR at feature level [19] are given in Table 2. The table shows that our method significantly improves the result over DWT with XOR at feature level fusion from 0.7052 to 0.7481.

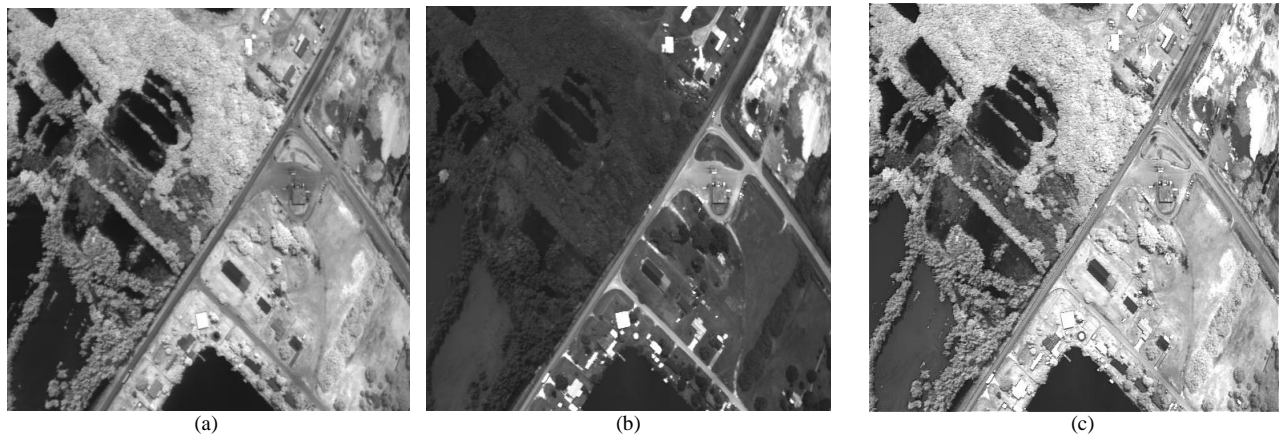

Fig.2 (a)-(b) are the input images, (c) is the corresponding fused image by using our method

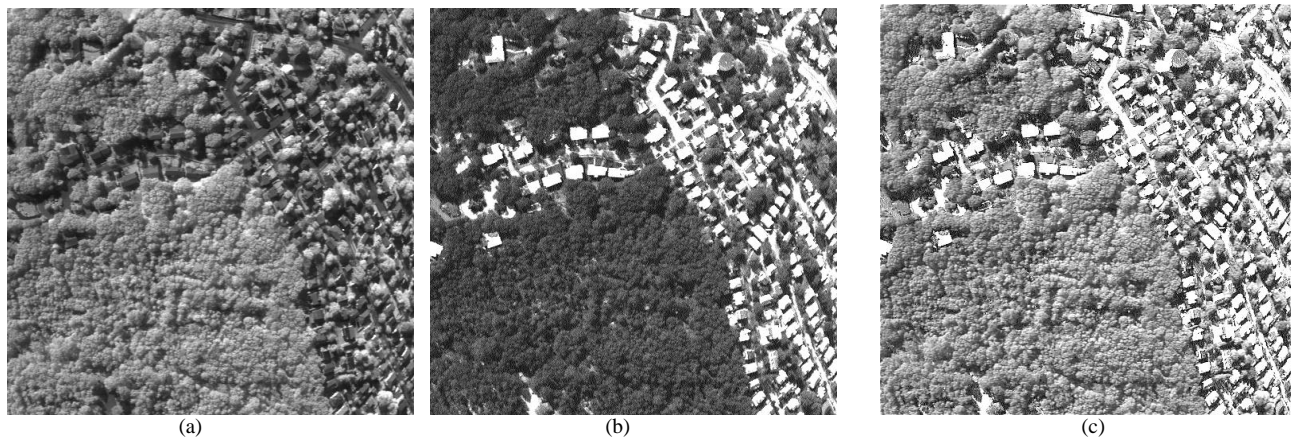

Fig.3 (a)-(b) are the input images, (c) is the corresponding fused image by using our method

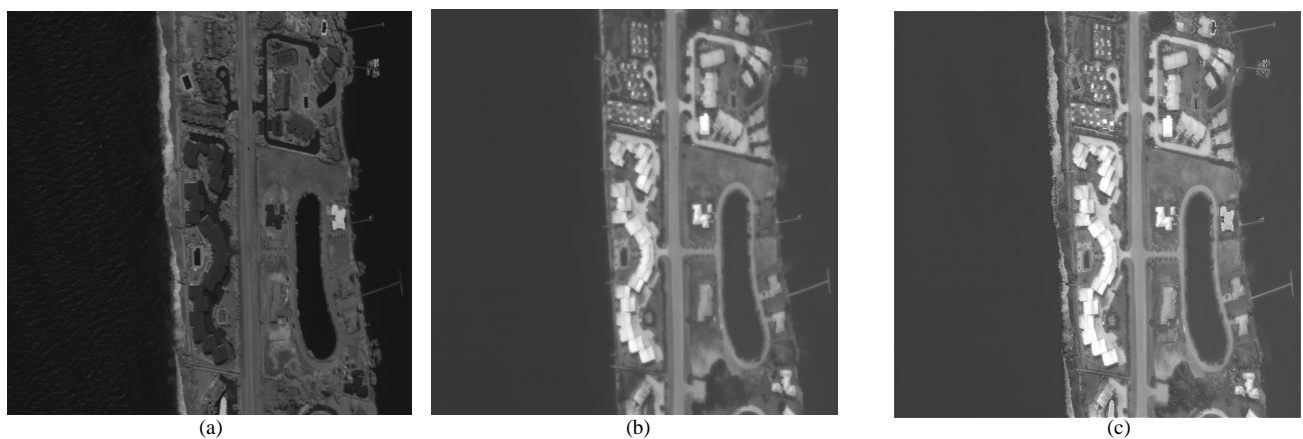

Fig.4 (a)-(b) are the input images, (c) is the corresponding fused image by using our method

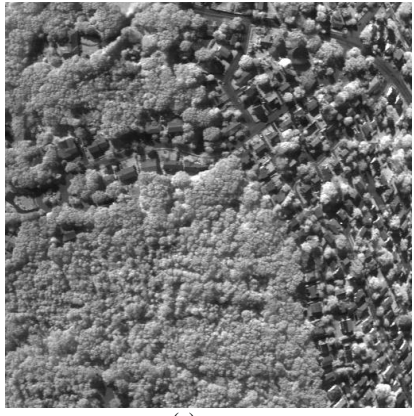
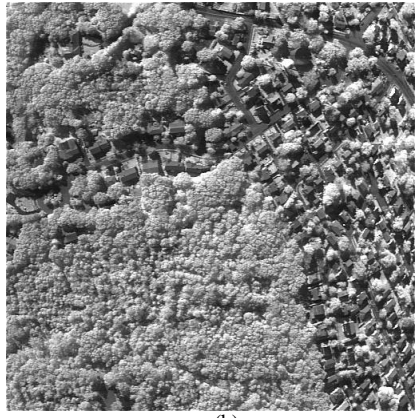

(a)                  (b)

Fig. 5 (a) is a input image, (b) filtered image after applying our filter mask

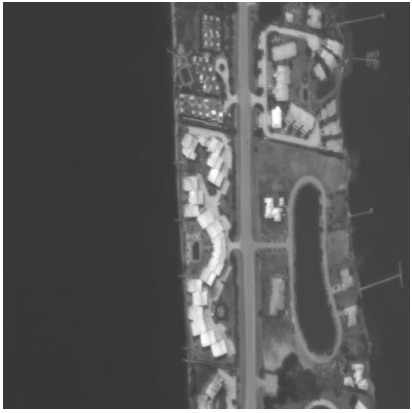
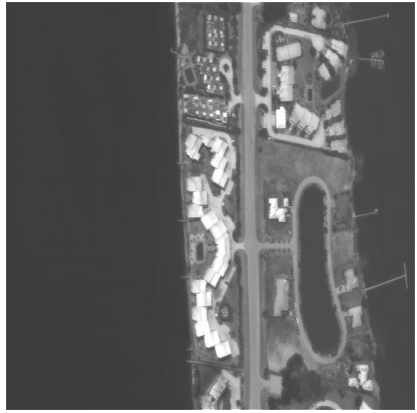

(a)                  (b)

Fig. 6 (a) is a input image, (b) filtered image after applying our filter mask

TABLE I. THE CUMULATIVE EVALUTION MEASURES OF FOUR IMAGE FUSION METHODS

| Method | | MIM | SD | E |
|---|---|---|---|---|
| Wavelet Transform | Fig.2 | 2.2482 | 63.7002 | 6.0241 |
| | Fig.3 | 2.5148 | 73.3618 | 6.7401 |
| | Fig.4 | 3.1120 | 60.0214 | 5.0013 |
| PCA | Fig.2 | 2.6422 | 78.7185 | 6.1012 |
| | Fig.3 | 2.6115 | 86.8222 | 6.5100 |
| | Fig.4 | 2.8802 | 70.3116 | 6.3202 |
| Weighted Average DWT using GA | Fig.2 | 3.3521 | 87.1206 | 8.1071 |
| | Fig.3 | 3.3005 | 90.0522 | 7.7013 |
| | Fig.4 | 3.7246 | 69.3730 | 7.0072 |
| Proposed Method | Fig.2 | 4.7021 | 167.5000 | 8.6896 |
| | Fig.3 | 4.5217 | 184.2889 | 8.2877 |
| | Fig.4 | 5.7382 | 100.6300 | 7.8555 |

TABLE II. AVERAGE $Q_P^{AB/F}$ SCORE OF TWO DIFFERENT METHODS

| Fusion scheme | $Q_P^{AB/F}$ Score |
|---|---|
| DWT+XOR at feature level fusion [19] | 0.7052 |
| Proposed Method | 0.7481 |

$$Q_P^{AB/F} = \frac{\sum_{n=1}^{N} \sum_{m=1}^{M} Q^{AF}(n,m) W^A(n,m) + Q^{BF}(n,m) W^B(n,m)}{\sum_{i=1}^{N} \sum_{j=1}^{M} (W^A(i,j) + W^B(i,j))} \quad \dots (3)$$

Where,

$A$, $B$ are the input images and $F$ is the fused image.

$Q^{AF}(n,m)$ and $Q^{BF}(n,m)$ are edge information preservation values. where, $0 \leq Q^{AF}(n,m) \leq 1$ and $0 \leq Q^{BF}(n,m) \leq 1$

$W^A(n,m)$ and $W^B(n,m)$ are weights of input image $A$ and $B$ at coordinate $(n,m)$

## IV. CONCLUSION

In this paper, we have proposed a fusion method based on the salient features. In our system, first we have applied a mask on two input images in order to conserve the high frequency information along with some low frequency information and stifle noise to a maximum extent. Thereafter, for identification of salience features from sources images, a local moment is computed in the neighborhood of a coefficient. Finally, a decision map is generated based on local moment in order to get the fused image. To validate this new approach, the approach was tested on 120 sensor image pairs collected from Manchester University UK database. Region based fusion using moment calculation may be employed in future for getting better fusion result.

ACKNOWLEDGMENT

Authors are thankful to Department of Computer Science and Engineering, Jadavpur University and National Institute of Science and Technology, India for providing necessary infrastructure to conduct experiments relation to this work.